\documentclass[12pt,final]{l4dc2021}

\PassOptionsToPackage{numbers, compress}{natbib}

\usepackage{enumitem}
\setlist[itemize]{noitemsep, topsep=0pt}

\usepackage{wrapfig}

\usepackage{xcolor}
\usepackage{colortbl}
\definecolor{linkcolor}{RGB}{74, 102, 146}
\definecolor{sacemph}{RGB}{168, 141, 201}
\usepackage{bbding}

\usepackage[misc]{ifsym}

\usepackage[utf8]{inputenc}
\usepackage[T1]{fontenc}
\usepackage[colorlinks=true,allcolors=linkcolor,pageanchor=true,plainpages=false,pdfpagelabels,bookmarks,bookmarksnumbered]{hyperref}
\usepackage{url}
\usepackage{booktabs}
\usepackage{amsfonts} %
\usepackage{nicefrac}
\usepackage{microtype}

\definecolor{cellhicolor}{RGB}{190, 215, 230}
\newcommand{\cellhi}{\cellcolor{cellhicolor}}

\usepackage{algorithm}

\usepackage{algpseudocode}
\algnewcommand{\LeftCommentX}[1]{\Statex \(\triangleright\) #1}
\algnewcommand{\LeftComment}[1]{\State \(\triangleright\) #1}
\makeatother

\usepackage{tikz}
\newcommand{\cblock}[3]{
  \hspace{-1.5mm}
  \begin{tikzpicture}
    [
    node/.style={square, minimum size=10mm, thick, line width=0pt},
    ]
    \node[fill={rgb,255:red,#1;green,#2;blue,#3}] () [] {};
  \end{tikzpicture}%
}

\usepackage{framed}
\usepackage{listings,textcomp,color}
\definecolor{backcolour}{rgb}{0.95,0.95,0.92}
\definecolor{deepblue}{rgb}{0,0,0.5}
\definecolor{deepred}{rgb}{0.6,0,0}
\usepackage[normalem]{ulem}

\usepackage{amsmath,amsfonts,bm,mathtools,amssymb}

\def\1{\bm{1}}

\def\gD{{\mathcal{D}}}

\def\gH{{\mathcal{H}}}

\def\gJ{{\mathcal{J}}}

\def\gN{{\mathcal{N}}}

\def\gQ{{\mathcal{Q}}}

\def\gU{{\mathcal{U}}}

\def\gX{{\mathcal{X}}}

\DeclareMathOperator*{\E}{\mathbb{E}}

\newcommand{\R}{\mathbb{R}}

\DeclarePairedDelimiterX{\infdivx}[2]{(}{)}{%
  #1\;\delimsize|\delimsize|\;#2%
}
\newcommand{\kl}[2]{\ensuremath{{\rm D}_{\rm KL}\infdivx{#1}{#2}}\xspace}

\newcommand{\defeq}{\vcentcolon=}

\newcommand{\gradupdate}{\ensuremath{\mathrm{grad\_update}}}

\newcommand{\svgH}{$\textrm{SVG}(H)$\xspace}
\newcommand{\sacsvgH}{$\textrm{SAC-SVG}(H)$\xspace}

\newcommand{\thispaper}{{{\color{gray} --- this paper}}}

\usepackage[nameinlink]{cleveref}
\Crefname{section}{Sect.}{Sects.}
\Crefname{appendix}{App.}{Apps.}
\Crefname{proposition}{Prop.}{Props.}

\usepackage{xspace}
\newcommand{\eg}{{\it e.g.}\xspace}
\newcommand{\ie}{{\it i.e.}\xspace}

\makeatletter
 \let\Ginclude@graphics\@org@Ginclude@graphics
\makeatother

\begin{document}
\title{On the model-based stochastic value gradient for
  \vbox{continuous reinforcement learning}}

\author{
  \Name{Brandon Amos}$^{1\;\textrm{\normalfont \Letter}}$\quad \Name{Samuel Stanton}$^2$\quad \Name{Denis Yarats}$^{1,2}$\quad
  \Name{Andrew Gordon Wilson}$^2$ \\
  \hspace*{0mm} \textnormal{$^1$Facebook AI Research \quad $^2$NYU \hfill
    $^\textrm{\Letter}$\hspace{0.5mm}Correspondence to: \texttt{bda@fb.com}}
}

\maketitle

\begin{abstract}%
For over a decade, model-based reinforcement learning has been seen
as a way to leverage control-based domain knowledge to improve the
sample-efficiency of reinforcement learning agents. While model-based
agents are conceptually appealing, their policies tend to lag behind
those of model-free agents in terms of final reward, especially in
non-trivial environments. In response, researchers have proposed
model-based agents with increasingly complex components, from
ensembles of probabilistic dynamics models, to heuristics for
mitigating model error. In a reversal of this trend, we show that
simple model-based agents can be derived from existing ideas that not
only match, but outperform state-of-the-art model-free agents in terms
of both sample-efficiency and final reward. We find that a model-free
soft value estimate for policy evaluation and a model-based stochastic
value gradient for policy improvement is an effective combination,
achieving state-of-the-art results on a high-dimensional humanoid
control task, which most model-based agents are unable to solve. Our
findings suggest that model-based policy evaluation deserves closer
attention.
\end{abstract}


\hspace{1mm}{\bgroup\leftskip 20pt\rightskip 20pt \small\noindent{\bfseries
Source Code:}}
\href{http://github.com/facebookresearch/svg}{github.com/facebookresearch/svg}

\section{Introduction}
The task of designing a reinforcement learning (RL) agent that can learn to optimally interact with an unknown environment has wide-reaching
applications in, \eg,
robotics \citep{kober2013reinforcement,polydoros2017survey},
control \citep{lillicrap2015continuous,kiumarsi2017optimal},
finance \citep{fischer2018reinforcement},
and gaming \citep{vinyals2019grandmaster,berner2019dota,justesen2019deep}.
Many long-standing challenges remain after decades of research
\citep{sutton2018reinforcement} and
the field is unsettled on a single method
for solving classes of tasks.
Subtle variations in the settings
can significantly impact how agents need to learn,
explore, and represent their internal decision process and the
external world around them.

\textit{Model-based} methods
explicitly construct a surrogate of the true environment, which can be
queried with hypothetical sequences of actions to assess their
outcome. In contrast, \textit{model-free} methods rely
entirely on online and historical ground-truth data, and implicitly
incorporate the environment dynamics into action value
estimates. Historically, the abstractions of model-based methods are
more amenable to the incorporation of expert domain knowledge
\citep{todorov2012mujoco}, and
often find higher reward policies than their model-free counterparts
when only a small amount of ground-truth data can be
collected \citep{deisenroth2011pilco, chua2018deep, wang2019exploring, janner2019trust, kaiser2019model}. However model-based methods struggle to match the
performance of model-free agents when the latter are allowed
unlimited interactions with the environment. While
recent work has greatly improved the asymptotic performance of
model-based agents, on the most challenging tasks they still often significantly under-perform model-free approaches.

While the incorporation of an explicit world model can introduce
helpful inductive biases to RL agents, imperfect learned
models introduce additional, unwanted biases.
An imperfect model may assign high expected reward to catastrophic
actions, or it may conjure fantasies of hypothetical states
having no correspondence to any realizable agent configuration.
Precisely which biases are introduced, and the resulting impact on the
agent, depends heavily on how the model is used.
In some cases, the model aids action
selection online, and in others it augments an existing
replay buffer with fictional experience.
The reproducibility crisis in reinforcement learning
\citep{islam2017reproducibility,henderson2018deep,engstrom2020implementation,sinha2020d2rl}
makes evaluating and understanding the tradeoffs difficult.

In this work we seek to answer a key question: \emph{``What is the
simplest model-based agent that can achieve state-of-the-art results
on current benchmark tasks?''} We show that if the model is used only
for policy improvement, then a simple combination of a stochastic
value gradient (SVG) \citep{heess2015learning}, entropy
regularization, soft value estimates \citep{haarnoja2018soft}, and a
single deterministic dynamics model are sufficient to match and
surpass the performance of model-based and model-free methods.
This finding is particularly notable because our baselines
are substantially stronger than those reported in previous work.
We also show that the performance improvements of our approach
cannot be attributed to model architecture alone.
Finally, we demonstrate that policy evaluation
can be significantly more sensitive to model bias than policy
improvement, even for short rollouts.

\begin{figure}[t]
  \centering
  \includegraphics[width=\textwidth]{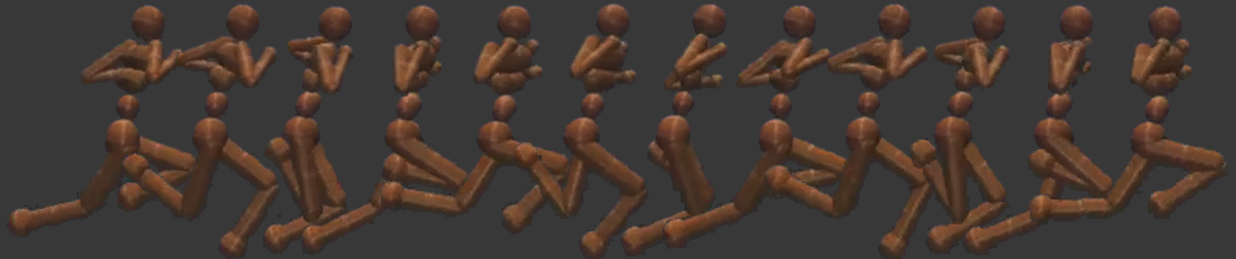}
  \caption{
    A model-based SAC-SVG agent learns a near-optimal humanoid gait
    by using short-horizon model rollouts.
    More videos are online at
    \href{http://sites.google.com/view/2020-svg}{sites.google.com/view/2020-svg}.
  }
  \label{fig:humanoid_gait}
\end{figure}

\section{Related work on model-based
  continuous reinforcement learning}
\label{sec:rw}

In most model-based methods for continuous control, the dynamics
model serves some or all of the following functions:
(1) a source of low-bias on-policy value estimates, obtained
by explicitly unrolling the policy for the full horizon or by refining
a bootstrapped value estimate; (2) a source of multistep value
gradients for policy improvement; or (3) a source of fictional
transition data with which to augment expensive ground-truth
transition data for standard model-free updates.
Nearly every combination of these three functions can be found
in the literature.
If bootstrapped value
estimates are not used, and the model fulfills functions (1) and (2), one
obtains a generalized form of PILCO \citep{deisenroth2011pilco}.
Introducing bootstrapped value estimates and removing (1), one obtains
the value gradient
\citep{heess2015learning,byravan2019imagined,hafner2019dream,clavera2020model}.
The model with (1) can also be used in a search procedure
\citep{springenberg2020local,marino2020iterative}.
Conversely if (1) is retained and (2) is removed, the MVE method
\citep{feinberg2018model} is recovered.
Dyna-Q \citep{sutton1990integrated} is an early method which uses
the model exclusively for (3), followed more recently by
NAF \citep{gu2016continuous}, ME-TRPO \citep{kurutach2018model},
and MBPO \citep{janner2019trust}.
Other contributions focus on the effect of ensembling some or
all of the learned models, such as
PETS \citep{chua2018deep}, STEVE \citep{buckman2018sample},
and POPLIN \citep{wang2019exploring}.
We overview the current state of the literature and summarize
some key dimensions in \cref{tab:rw}.
Our paper fits into this space of related work to show that
if set up correctly, the value gradient is a competitive baseline.

We have focused this section on continuous spaces and
refer to \citet{schrittwieser2020mastering,hamrick2020role}
for further discussions and related work in discrete spaces.

\begin{table*}[!t]
  \caption{
    Key differences between related work on imagination,
    value-expansion, and policy distillation for \emph{continuous control}.
    We use \emph{MVE} to denote \emph{some} form of value-expansion,
    which may not necessarily have an explicit terminal value
    approximation.
  }
  \resizebox{1\textwidth}{!}{\begin{tabular}{l|cc|c|ccc}
\toprule
& \multicolumn{2}{c|}{Policy Learning} & Value Learning & \multicolumn{3}{c}{Dynamics} \\
& Update & Objective & & Model & Ensemble & Observation Space \\ \midrule
PILCO \citep{deisenroth2011pilco} & G & MBPG & - & GP & No & Proprio \\
MVE \citep{feinberg2018model} & G & MF & MVE & Det NN & No & Proprio \\
STEVE \citep{buckman2018sample} & G & MF & MVE & Prob NN & Yes & Proprio \\
IVG \citep{byravan2019imagined} & G & MVE & MF & Det NN & No & Pixel+Proprio \\
Dreamer \citep{hafner2019dream} & G & MVE & MVE & Prob NN & No & Pixel \\ \midrule
GPS \citep{levine2013guided} & BC & MVE & - & Local & No & Proprio \\
POPLIN \citep{wang2019exploring} & BC & MVE & - & Prob NN & Yes & Proprio \\ \midrule
METRPO \citep{kurutach2018model} & G & MF+rollout data & MF+rollout data & Det NN & Yes & Proprio \\
MBPO \citep{janner2019trust} & G & MF+rollout data & MF+rollout data & Prob NN & Yes & Proprio \\ \midrule
SAC \citep{haarnoja2018soft} & G & MF & MF &   & - \\
\textbf{SAC-SVG(H)} \thispaper & G & MVE & MF & Det NN & No & Proprio \\
\bottomrule
\end{tabular}}
\resizebox{1\textwidth}{!}{
  G=Gradient-based \hfill
  BC=Behavioral Cloning \hfill
  MF=Model Free \hfill
  MVE=Model-Based Value Expansion \hfill
  GP=Gaussian Process
}

  \label{tab:rw}
\end{table*}

\section{Preliminaries, notation, and background in
reinforcement learning}
\label{sec:background}
Here we present the
non-discounted setting for brevity and refer to
\citet{thomas2014bias,haarnoja2018soft}
for the full details behind the
$\gamma$-discounted setting.

\subsection{Markov decision processes and reinforcement learning}
A Markov decision process (MDP) \citep{szepesvari2010algorithms,puterman2014markov}
is a discrete-time stochastic control process widely used
in robotics and industrial systems.
We represent an MDP as the tuple
$(\gX, \gU, p, r)$,
where $\gX$ is the \emph{state space},
$\gU$ is the \emph{control} or \emph{action space},
the \emph{transition dynamics} $p\defeq\Pr(x_{t+1}|x_t,u_t)$
capture the distribution over the next state $x_{t+1}$
given the current state $x_t$ and control $u_t$,
$r: \gX\times\gU\rightarrow\R$ is the \emph{reward map}.
The \emph{termination map} $d: \gX\times \gU\rightarrow[0,1]$
indicates the probability of the system
terminating after executing $(x_t, u_t)$.
We refer to the dynamics, rewards, and termination map as
the \emph{world model}, \eg following \citet{ha2018worldmodels},
and consider MDPs with known
\emph{and} unknown world models.
We focus on MDPs with \emph{continuous} and
\emph{real-valued} state spaces $\gX=\R^m$ and
\emph{bounded} control spaces $\gU=[-1,1]^n$.
We consider parameterized \emph{stochastic policies}
$\pi_\theta(x_t) \defeq \Pr(u_t|x_t)$
that induce \emph{state} and \emph{state-control marginals}
$\rho_t^\pi(\cdot)$ for each time step $t$, which can be constrained
to start from an initial state $x_0$ as $\rho_t^\pi(\cdot | x_0)$.
For finite-horizon non-discounted MDPs, the \emph{value} $V$
or \emph{action-conditional value} $Q$
of a policy $\pi$ is
\begin{equation}
  V^\pi(x)\defeq\sum_t\E_{(x_t, u_t)\sim\rho_t^\pi(\cdot|x)} r(x_t, u_t), \qquad
  Q^\pi(x,u) \defeq r(x,u) + \E_{x'\sim \rho_1^\pi(\cdot|x)} V^\pi(x'),
  \label{eq:value}
\end{equation}
which may be extended to regularize the policy's entropy with
some temperature $\alpha\geq 0$ as
\begin{equation}
  V^{\pi,\alpha}(x)\defeq\sum_t\E_{(x_t, u_t)\sim\rho_t^\pi(\cdot|x)}
      r(x_t, u_t) - \alpha \log \pi(u_t|x_t)
  \label{eq:value-entr}.
\end{equation}
The value function of a given policy can be estimated
(\ie \emph{policy evaluation}) by explicitly rolling out the
world model for $H$ steps with
\begin{equation}
  \begin{split}
    V^\pi_{0:H}(x)&\defeq \sum_{t< H}\E_{(x_t, u_t)\sim\rho_t^\pi(\cdot|x)} r(x_t, u_t) +
    \E_{x_H\sim\rho_H^\pi(\cdot|x)} \tilde V(x_H), \\
  Q^\pi_{0:H}(x,u)&\defeq r(x,u) + \E_{x'\sim\rho_1^\pi(\cdot|x)} V^\pi_{0:H-1}(x').
  \end{split}
  \label{eq:mve}
\end{equation}
The final value estimator $\tilde V$ can take the form of a simple
terminal reward function or a parametric function approximator trained
through some variant of Bellman backup such as fitted Q-iteration
\citep{antos2008fitted}.
Following \citet{feinberg2018model}, we refer to \cref{eq:mve}
as the \emph{model-based value expansion} (MVE).

\emph{Policy improvement} updates the policy to attain a
better expected value.
In the RL setting, the world model
is often unknown and the value estimate is approximated in
a \emph{model-free} way that does not attempt to
explicitly model the world.
In many \emph{actor-critic} methods, policy improvement
is done with gradient ascent using the
\emph{value gradient} $\nabla_\theta V^\pi(x)$.
With stochastic policies, $\nabla_\theta V^\pi(x)$ is the
\emph{stochastic value gradient} (SVG) \citep{heess2015learning}.
For consistency, we refer to methods that update the policy with an $H$-step
value expansion as \svgH, even though other papers refer to this
by other names
\citep{byravan2019imagined,hafner2019dream,clavera2020model}.

\subsection{The soft actor-critic for learning
  continuous control policies}
\label{sec:bg:sac}
The \emph{soft actor-critic} (SAC) method \citep{haarnoja2018soft}
learns a \emph{state-action value function} $Q_\theta$,
\emph{stochastic policy} $\pi_\theta$, and a \emph{temperature} $\alpha$
to find an optimal policy for a continuous-valued MDP
$(\gX, \gU, p, r, \gamma)$.
SAC optimizes a $\gamma$-discounted \emph{maximum-entropy objective} as in
\citet{ziebart2008maximum,ziebart2010modeling,fox2015taming}.
The \emph{policy} $\pi_\theta$ is a parameterized $\tanh$-Gaussian that
given $x_t$, generates
samples
$u_t = \tanh(\mu_\theta(x_t)+\sigma_\theta(x_t)\epsilon)$,
where $\epsilon\sim\gN(0,1)$
and $\mu_\theta$ and $\sigma_\theta$ are models
that generate the pre-squashed mean and standard deviation.
The \emph{critic} $Q_\theta$ optimizes the
\emph{soft (single-step) Bellman residual}
\begin{equation}
  \small
  \gJ_Q(\gD) \defeq \E_{(x_t,u_t)\sim\gD}[
    (
      Q_\theta(x_t, u_t) - Q_{\bar\theta}^{\rm targ}(x_t, u_t)
    )^2
  ],
\label{eq:sac:soft_bellman}
\end{equation}
where $\gD$ is a distribution of transitions, \eg an offline
\emph{replay buffer} containing recent experience,
$\bar\theta$ is an exponential moving average of the weights
\citep{mnih2015human},
\begin{equation}
Q_{\bar\theta}^{\rm targ}(x_t, u_t) \defeq
r(x_t,u_t) + \gamma\E_{x_{t+1}\sim p} V_{\bar\theta}(x_{t+1}),
\end{equation}
is the critic target, the \emph{soft value function} is
\begin{equation}
V_{\bar\theta}(x) \defeq \E_{u\sim\pi_\theta(x)}\left[
    Q_{\bar\theta}(x, u)-\alpha \log \pi_\theta(u|x)\right],
\end{equation}
and
$\log\pi_\theta(u|x)$ is the log-probability of the action.
SAC also uses \emph{double Q learning} \citep{hasselt2010double}
and does policy optimization with the objective
\begin{equation}
  \gJ_{\pi,\alpha}^{\rm SAC}(\gD) \defeq \E_{x\sim\gD}\left[
    \kl{\pi_\theta(\cdot|x)}{\gQ_{\theta,\alpha}(x, \cdot)}
  \right]
  = \E_{x\sim\gD,u\sim\pi(x)}\left[\alpha \log \pi(u|x) - Q(x,u) \right],
  \label{eq:sac:policy}
\end{equation}
where
$\gQ_{\theta,\alpha}(x, \cdot) \propto \exp\{\frac{1}{\alpha} Q_{\theta,\alpha}(x, \cdot)\}$
and the last equality comes from expanding the KL definition.
The temperature $\alpha$ is adapted following \citet{haarnoja2018soft}
to make the policy's entropy
match a target value $\bar\gH\in\R$ by optimizing
\begin{equation}
  \gJ_\alpha(\gD) \defeq \E_{x_t\sim\gD,u_t\sim\pi_\theta(x_t)}\left[
    -\alpha \log\pi_\theta(u_t|x_t) - \alpha \bar\gH
  \right].
  \label{eq:sac:temp}
\end{equation}

\section{\svgH with entropy regularization and
  a model-free value estimate}

In this section we discuss a simple model-based extension of model-free SAC. We find that one is immediately suggested by reframing the SAC actor update within the SVG framework. We also describe our dynamics model architecture, which is better suited for multistep value gradients than the MLPs often employed in model-based RL agents.

\subsection{Connecting the SAC actor update and
  stochastic value gradients}

Although motivated differently in
\citet{haarnoja2018soft}, the SAC actor update is
equivalent to entropy-regularized SVG(0) with a soft value estimate.
This is seen by comparing the entropy-regularized value estimate in
\cref{eq:value-entr} with the SAC actor objective in
\cref{eq:sac:policy}.

Given the empirical success of model-free SAC agents, it is natural to
think about a model-based extension. Since the model-free SAC actor
update is entropy-regularized SVG(0), simply using the same soft value
estimate and entropy adjustment heuristic from SAC, and increasing the
SVG horizon from 0 to H immediately provides such an
extension. The approach retains some of the desirable characteristics
of SAC, such as the effectiveness of the soft value estimate in
encouraging exploration, but adds the ability of multi-step SVG to use
on-policy value expansions for policy improvement.
\Cref{alg:sac-svg} summarizes the approach, which we will refer to
\sacsvgH, with more details provided in \cref{app:exp}. Briefly
put, for policy improvement we minimize the entropy-regularized value
estimate
\begin{equation}
  \gJ_{\pi,\alpha}^{\rm SVG}(\gD) \defeq \E_{x\sim\gD} -V^{\pi,\alpha}_{0:H}(x)
  \label{eq:svg_policy}
\end{equation}
by ascending the stochastic value gradient, differentiating \cref{eq:svg_policy} with respect
to $\theta$. $V^{\pi,\alpha}_{0:H}$ is an entropy-regularized value expansion
that explicitly unrolls the model
for H steps and uses the model-free SAC value estimate at the terminal rollout state-action pair.
We adapt the temperature $\alpha$ to make the
policy's expected entropy match a target value $\bar\gH\in\R$
by optimizing the $\gJ_\alpha$ from SAC in \cref{eq:sac:temp}.

\newpage
\subsection{Approximate world models for deterministic systems}
\begin{wrapfigure}{r}{0.37\textwidth}
  \centering
  \vspace{-6mm}
  \includegraphics[width=0.37\textwidth]{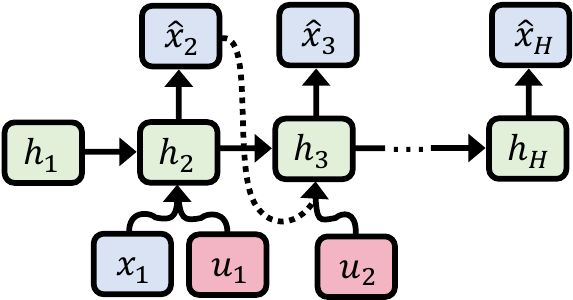}
  \vspace{-6mm}
\end{wrapfigure}
We learn a \emph{deterministic transition dynamics model}
$f_\theta$ that maps the current state
and a sequence of actions to the next sequence
of states, \ie $f_\theta:\gX\times\gU^{H-1}\rightarrow\gX^{H-1}$.
Our dynamics $f_\theta$ is autoregressive
over time and predicts state deltas with
a GRU~\citep{cho2014learning} carrying
forward a hidden state.
We start with the current system state $x_1$ and an initial
hidden state $h_1$ and use $h_1=0$ in all of our experiments.
Given $\{x_t, h_t, u_t\}$, we encode the state and action
into $z_t\defeq f^{\rm enc}_\theta(x_t, u_t)$. We then use a
multi-layer GRU to update the hidden state with
$h_{t+1}\defeq {\rm GRU}(z_t, h_t)$, and we then decode the hidden
state with $\hat x_{t+1}\defeq x_t + f^{\rm dec}_\theta(h_{t+1})$.
We model $f^{\rm enc}_\theta$ and $f^{\rm dec}_\theta$
with multi-layer MLPs.
\Cref{app:exp} describes more experimental details and our
hyper-parameters. For every task we consider, we use a
2-layer MLPs with 512 hidden units and 2-layer GRU.

Learning the dynamics uses multi-step squared error loss
\begin{equation}
  \gJ_f(\gD) \defeq \E_{(x_{1:H},u_{1:H-1})\sim\gD}\left[
    ||f_\theta(x_1, u_{1:H-1}) - x_{2:H}||_2^2
  \right],
  \label{sec:mve_sac:Jf}
\end{equation}
where $f_\theta(x_1, u_{1:H-1})$ unrolls the GRU and predicts
the next states $\hat x_{2:H}$ and the expectation
$\E_{(x_{1:H},u_{1:H-1})\sim\gD}$ is over $H$-step sequences
uniformly from the replay buffer.
We approximate all expectations for the models in this section
with a batch size of 1024.

We learn the \emph{reward map}
$r_\theta:\gX\times\gU\rightarrow \R$
as a neural network
with a squared error loss
\begin{equation}
  \gJ_r(\gD) \defeq \E_{(x_t,u_t,r_t)\sim\gD}\left[
    ||r_\theta(x_t, u_t)-r_t||_2^2
  \right].
  \label{sec:mve_sac:Jr}
\end{equation}

We learn the \emph{termination map}
$d_\theta: \gX\times \gU\rightarrow [0, 1]$
that predicts the probability of the system
terminating after executing $(x_t, u_t)$,
which we observe as $d_{t+1}\in\{0,1\}$.
We model $d_{t+1}$ as a Bernoulli response with likelihood
\begin{equation}
  \Pr(d_{t+1}|x_t,u_t) \defeq
    d_\theta(x_t,u_t)^{d_{t+1}}(1-d_\theta(x_t, u_t))^{1-d_{t+1}}.
  \label{eq:bernoulli-term}
\end{equation}
We use a multi-layer neural network to
model $d_\theta$ in the logit space
and minimize the negative log-likelihood
with the objective
\begin{equation}
  \gJ_d(\gD) \defeq \E_{(x_t,u_t,d_{t+1})\sim\gD}\left[
    -\log d_\theta(d_{t+1}|x_t, u_t) \right].
  \label{sec:mve_sac:Jd}
\end{equation}
We only learn the terminations that are not time-based.
\Cref{sec:mve_sac:Jd} could be weighted to deal with an
imbalance between termination and non-termination conditions,
but in practice we found this weighting to not be important.

\textbf{Discussion.}
Our deterministic dynamics model using a GRU for predicting
multi-step dynamics is simple in comparison to the ensembles
of probabilistic models other recent methods use
\citep{chua2018deep,buckman2018sample,janner2019trust}.
The improvements from these models could benefit our
base model as well, but our design choice here for simplicity
enables us to
1) sample states uniformly from the replay buffer
without needing to obtain the most recent hidden
state associated with it,
2) still model transitions in a latent space of the GRU,
and
3) optimize the multi-step likelihood.

\newpage
\section{Experimental results on MuJoCo locomotion control tasks\protect\footnote{
Our source code is online at
\href{http://github.com/facebookresearch/svg}{github.com/facebookresearch/svg} and
builds on the SAC implementation from
\citet{yarats2020pytorch_sac}.
Videos of our agents are available at
\href{http://sites.google.com/view/2020-svg}{sites.google.com/view/2020-svg}.
}}

We evaluate \sacsvgH on \emph{all} of the
MuJoCo \citep{todorov2012mujoco} locomotion
experiments considered by POPLIN, MBPO, and STEVE,
all current state-of-the-art
related approaches. Note that although we compare against those methods, elements of each could be introduced to \sacsvgH and improve the performance at the cost of increased complexity.
We provide a sweep over horizon lengths for the POPLIN tasks
and fix $H=2$ in the MBPO tasks.
For every horizon length, we perform a hyper-parameter search
only over the target entropy schedule, which we further
describe in \cref{app:exp}.
Our SAC baseline uses the same state normalization and
target entropy schedule as our SAC-MVE runs in every environment.

\Cref{tab:exp:poplin} shows the results of our method in comparison
to POPLIN \citep{wang2019exploring} on the locomotion tasks
they consider from \citet{wang2019benchmarkingmbrl},
POPLIN uses the ground-truth reward for these tasks and
learns a model --- we learn both.
We outperform POPLIN in most of the locomotion tasks, though it has a strong exploration strategy and works
exceedingly well in the cheetah environment
from PETS \citep{chua2018deep}.
Notably our SAC baseline often also outperforms the SAC baseline
reported in \citet{wang2019exploring}.
We are able to find a policy that generates an optimal
action with a \emph{single} rollout sample rather than
the \emph{thousands} of rollouts POPLIN typically uses
and find that setting $H=2$
usually improves upon our SAC baseline ($H=0$). \Cref{fig:exp:mbpo} shows our results in comparison to
MBPO and STEVE, which evaluate on the MuJoCo tasks
in the OpenAI gym \citep{brockman2016openai} that are
mostly the standard \verb!v2! tasks with early termination
and alive bonuses, and with a truncated observation space for
the humanoid and ant that discards the inertial measurement units.
\sacsvgH consistently matches the best performance
and convergence rates across every task and is
able to learn a running gait on the humanoid (\cref{fig:humanoid_gait}).

\begin{table*}[t!]
  \caption{
    \sacsvgH excels in the locomotion tasks
    considered in \citet{wang2019exploring}.
    We report the mean evaluation rewards and standard
    deviations across ten trials.
  }
  \hspace{-8mm}
\scalebox{0.65}{
\hspace{-4mm}
\rotatebox{90}{{{\color{gray} \hspace{6mm} this paper}}}
\hspace{-2mm}
\begin{tabular}{l|c|c|c|c|c||c}
\toprule
& Ant            & Hopper       & Swimmer      & Cheetah   & Walker2d & PETS Cheetah       \\
\midrule
SAC-SVG(1) & \cellhi 3691.00 $\pm$ 1096.77 & 1594.43 $\pm$ 1689.01 & \cellhi 348.40 $\pm$ 8.32 & 6890.20 $\pm$ 1860.49 & -291.54 $\pm$ 659.52 & 5321.23 $\pm$ 1507.00 \\
SAC-SVG(2) & \cellhi 4473.36 $\pm$ 893.44 & \cellhi 2851.90 $\pm$ 361.07 & \cellhi 350.22 $\pm$ 3.63 & \cellhi 8751.76 $\pm$ 1785.66 & 447.68 $\pm$ 1139.51 & 5799.59 $\pm$ 1266.93 \\
SAC-SVG(3) & \cellhi 3833.12 $\pm$ 1418.15 & 2024.43 $\pm$ 1981.51 & 340.75 $\pm$ 13.46 & \cellhi 9220.39 $\pm$ 1431.77 & 877.77 $\pm$ 1533.08 & 5636.93 $\pm$ 2117.52 \\
SAC-SVG(4) & 2896.77 $\pm$ 1444.40 & 2062.16 $\pm$ 1245.33 & 348.03 $\pm$ 6.35 & \cellhi 8175.29 $\pm$ 3226.04 & \cellhi 1852.18 $\pm$ 967.61 & 5807.69 $\pm$ 1008.60 \\
SAC-SVG(5) & 3221.66 $\pm$ 1576.25 & 608.58 $\pm$ 2105.60 & 340.99 $\pm$ 4.58 & 6129.02 $\pm$ 3519.98 & \cellhi 1309.20 $\pm$ 1281.76 & 4896.22 $\pm$ 1033.33 \\
SAC-SVG(10) & 1389.30 $\pm$ 981.59 & -2511.05 $\pm$ 881.26 & 303.16 $\pm$ 10.57 & 2477.25 $\pm$ 2596.43 & -2328.08 $\pm$ 735.48 & 4248.25 $\pm$ 802.54 \\ \hline
POPLIN-P~{\footnotesize \citep{wang2019exploring}} & 2330.1 $\pm$ 320.9 & 2055.2 $\pm$ 613.8 & 334.4 $\pm$ 34.2   & 4235.0 $\pm$ 1133.0 & 597.0 $\pm$ 478.8 & \cellhi 12227.9 $\pm$ 5652.8 \\ \hline
SAC*~{\footnotesize \citep{haarnoja2018soft}}      &
548.1 $\pm$ 146.6  &
788.3 $\pm$ 738.2  &
204.6 $\pm$ 69.3   &
3459.8 $\pm$ 1326.6 &
164.5 $\pm$ 1318.6  &
1745.9 $\pm$ 839.2   \\
SAC (our run) & 510.56 $\pm$ 76.38 & 2180.33 $\pm$ 977.30 & \cellhi 351.24 $\pm$ 5.27 & 6514.83 $\pm$ 1100.61 & \cellhi 1265.13 $\pm$ 1317.00 & 3259.99 $\pm$ 1219.94 \\ \hline
PETS*~{\footnotesize \citep{chua2018deep}}    &
1165.5 $\pm$ 226.9 &
114.9 $\pm$ 621.0  &
326.2 $\pm$ 12.6   &
2288.4 $\pm$ 1019.0 &
282.5 $\pm$ 501.6   &
4204.5 $\pm$ 789.0   \\
METRPO*~{\footnotesize \citep{kurutach2018model}}   &
282.2 $\pm$ 18.0   &
1272.5 $\pm$ 500.9 &
225.5 $\pm$ 104.6  &
2283.7 $\pm$ 900.4  &
-1609.3 $\pm$ 657.5 &
-744.8 $\pm$ 707.1   \\
TD3*~{\footnotesize \citep{fujimoto2018td3}}      &
870.1 $\pm$ 283.8  &
1816.6 $\pm$ 994.8 &
72.1 $\pm$ 130.9   &
3015.7 $\pm$ 969.8  &
-516.4 $\pm$ 812.2  &
218.9 $\pm$ 593.3    \\
                             \midrule
Training Timesteps &
200000 &
200000 &
50000 &
200000 &
200000 &
50000 \\
\bottomrule
\end{tabular} \\
}
\hspace*{-6.5mm} \scalebox{0.63}{
  *Denotes the baseline results reported in \citet{wang2019exploring}.}

  \label{tab:exp:poplin}
\end{table*}

\begin{figure*}[t!]
  \includegraphics[width=0.32\textwidth]{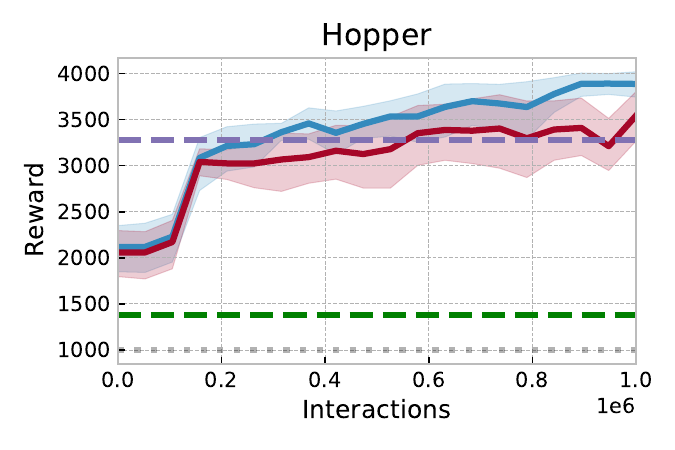}
  \includegraphics[width=0.32\textwidth]{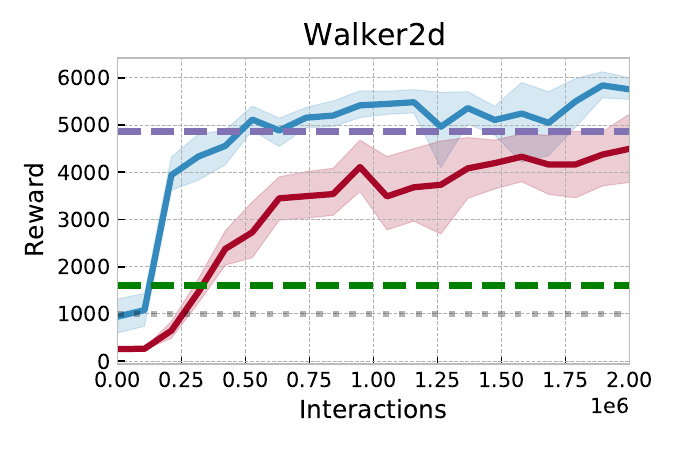}
  \includegraphics[width=0.32\textwidth]{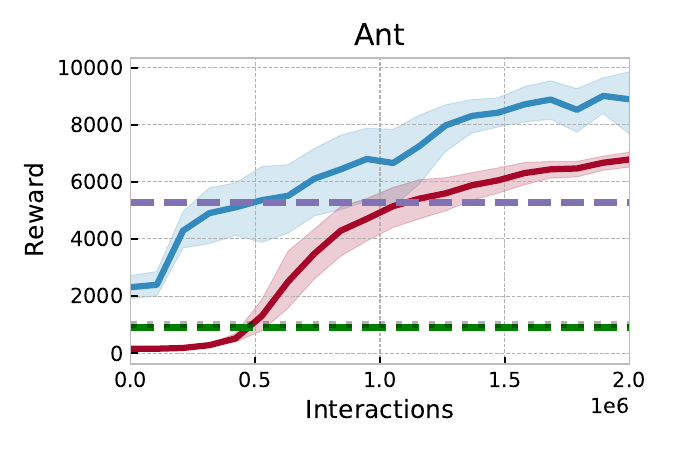} \\
  \includegraphics[width=0.32\textwidth]{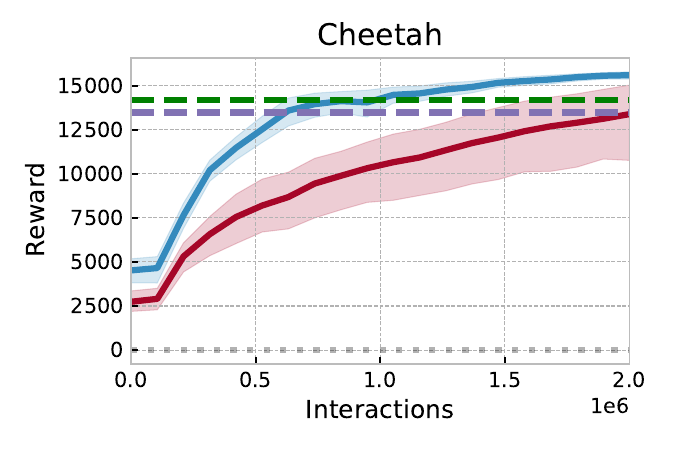}
  \includegraphics[width=0.32\textwidth]{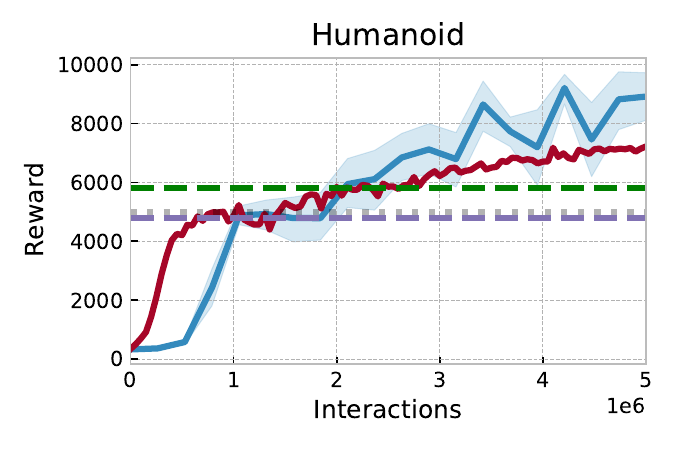}
  \hspace{4mm}
  \begin{tikzpicture}[every node/.style = {anchor=west,font=\footnotesize}]
    \node at (0,0) {\hspace{0.1mm}\cblock{52}{138}{189}\hspace{2.5mm} SAC-SVG(2) \thispaper};
    \node at (0,-4mm) {\hspace{0.1mm}\cblock{166}{6}{40}\hspace{2.5mm} SAC};
    \node at (0,-8mm) {\hspace{0.1mm}\cblock{128}{114}{179}\hspace{2.5mm} MBPO};
    \node at (0,-12mm) {\hspace{0.1mm}\cblock{6}{115}{0}\hspace{2.5mm} STEVE};
    \draw [line width=0.5mm,dotted,color={rgb,255:red,166;green,166;blue,166}] (0mm,-16mm) -- (5.5mm,-16mm);
    \node at (6mm,-16mm) {Alive bonus for surviving};
    \draw [line width=0.5mm,dashed] (0mm,-20mm) -- (6mm,-20mm);
    \node at (6mm,-20mm) {Convergence};
    \node at (0,-23mm) {}; %
  \end{tikzpicture}\vspace{-4mm}
  \caption{
    \sacsvgH excels in the locomotion tasks
    considered in \citet{janner2019trust}.
    We report the mean evaluation rewards and standard
    deviations across ten trials.
  }
  \label{fig:exp:mbpo}
\end{figure*}
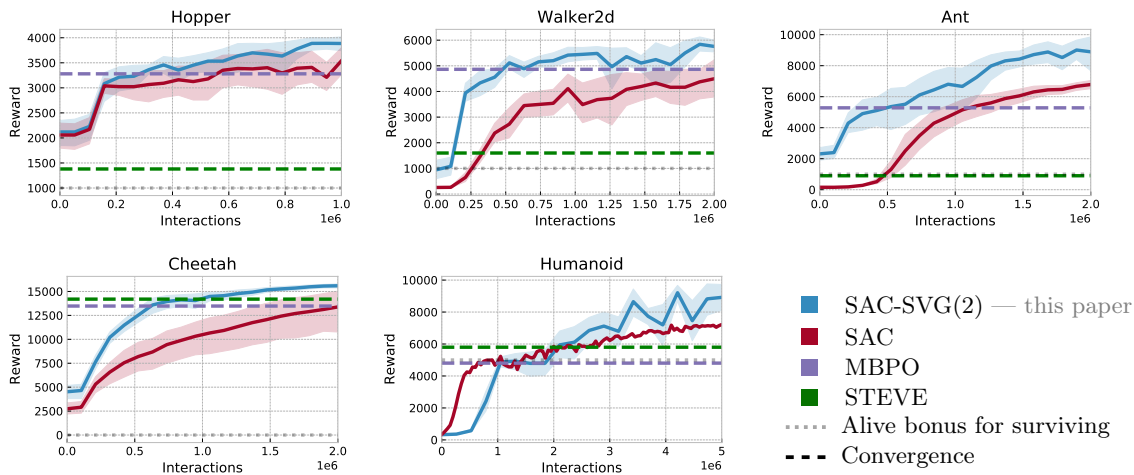

\subsection{Ablations}
\subsubsection{Model architectures and ensembling}

PETS, POPLIN, and MBPO all relied on very similar implementations of
bootstrapped ensembles of MLPs to model
dynamics. Since we use a different architecture, it is
important to consider the impact of the change on the agent's performance. Simply comparing the
reward curves resulting from different model choices does not provide any
insight into why some
methods perform better than others. In particular, such
head-to-head comparisons cannot differentiate between a model's
overall ability to generalize and more subtle interactions with the
rest of the agent, such as the effect on exploration.

\begin{figure*}[t!]
\begin{minipage}{0.32\textwidth}
\centering
\includegraphics[width=\textwidth]{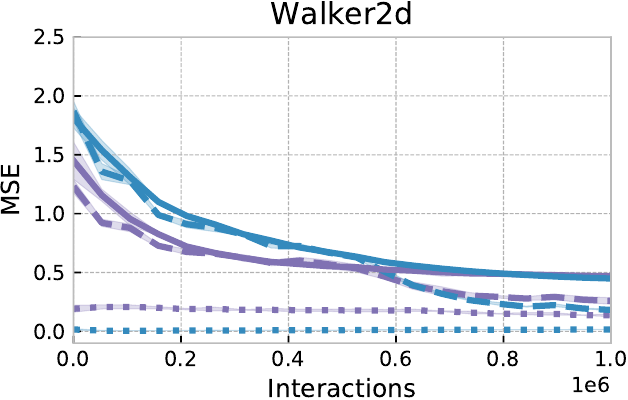}
\end{minipage}
\hfill
\begin{minipage}{0.32\textwidth}
\centering
\includegraphics[width=\textwidth]{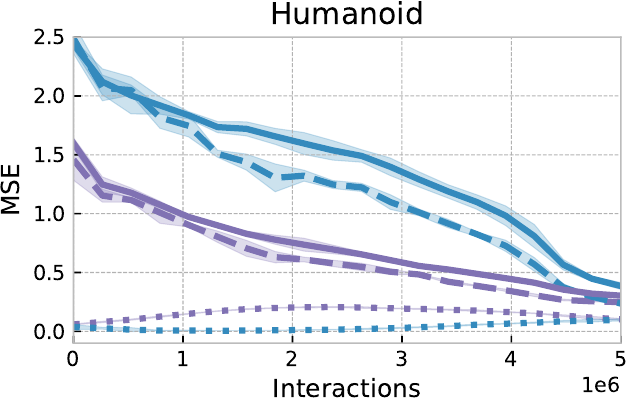}
\end{minipage}
\hfill
\begin{minipage}{0.33\textwidth}
  \begin{tikzpicture}[every node/.style = {anchor=west,font=\footnotesize}]
    \node at (0,-4mm) {\hspace{0.1mm}\cblock{52}{138}{189}\hspace{2.5mm} Recurrent Network \thispaper};
    \node at (0,-8mm) {\hspace{0.1mm}\cblock{128}{114}{179}\hspace{2.5mm} FCNet Ensemble};
    \draw [line width=0.5mm,dotted,color={rgb,255:red,166;green,166;blue,166}] (0mm,-12mm) -- (5.5mm,-12mm);
    \node at (6mm,-12mm) {Train Loss};
    \draw [line width=0.5mm,dashed] (0mm,-16mm) -- (6mm,-16mm);
    \node at (6mm,-16mm) {Holdout Loss};
    \draw [line width=0.5mm] (0mm,-20mm) -- (6mm,-20mm);
    \node at (6mm,-20mm) {Test Loss};
    \node at (0,-23mm) {}; %
  \end{tikzpicture}\vspace{-4mm}
\end{minipage}
\caption{
  We compare the multi-step MSE of a 5-component ensemble of MLPs trained on one-step transitions to that of a single recurrent
  network, trained on multi-step transition sequences. The gap between the
  holdout curve and the test curve is error due to distribution shift
  between the train and test
  distributions. Since the ensemble of MLPs
  generalizes better in the supervised setting, our performance
  improvement cannot be attributed to a simple change in model
  architecture.}
\label{fig:dx_architecture_ablation}
\end{figure*}
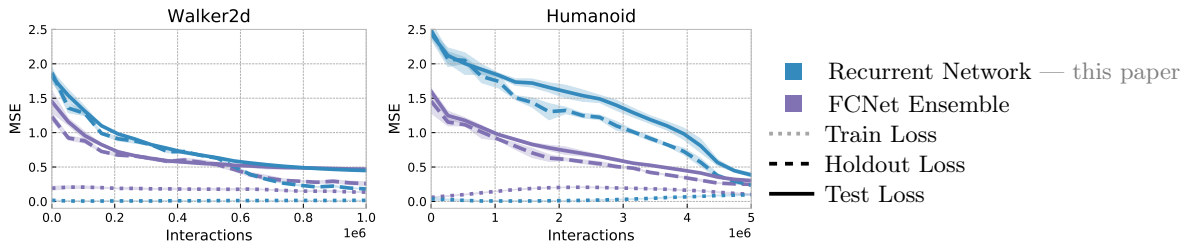

To explore the difference in generalization ability between different
architectures, we removed the dependence of the data collection
process on the dynamics model by comparing the architectures on an independent
episodic replay buffer, generated by a standard model-free SAC
trial.
By sequentially adding episodes to the model's
training and holdout datasets, retraining the model, and testing on a
fixed test split, we isolated the generalization characteristics of
the models
while simulating the online nature of the RL data-collection process. The
results of this experiment are presented
in \cref{fig:dx_architecture_ablation}.
An ensemble of MLPs generalizes better than a
single recurrent network of similar capacity. We selected the
recurrent network for its amenability to multistep value gradient
backpropagation. While we could ensemble our recurrent
architecture, we observe that a single model
obtains competitive results, and have chosen to prioritize simplicity.
Interestingly, even though the single
recurrent model overfits much more heavily to the training data, the
asymptotic reward of our humanoid agent is significantly higher and
qualitatively different than the agent in \citet{janner2019trust}.
Hence it is not clear how well the model \textit{needs} to generalize,
since short-horizon rollouts are typically initiated with
states from the replay buffer (\ie the model's training data).

\subsubsection{Value expansions in the actor and critic}

\begin{figure*}[t]
  \includegraphics[width=0.32\textwidth]{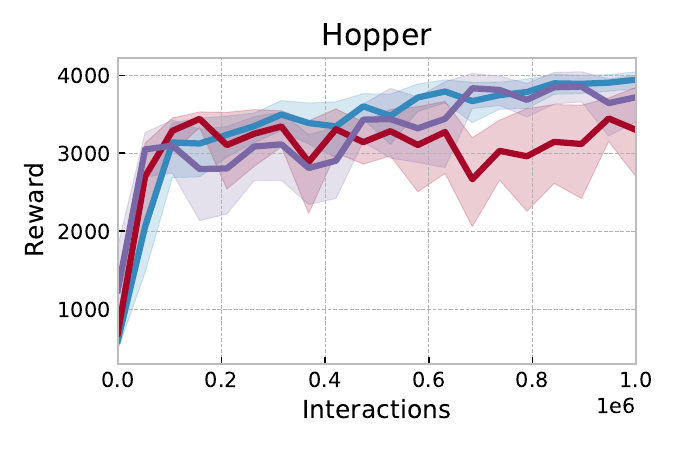}
  \includegraphics[width=0.32\textwidth]{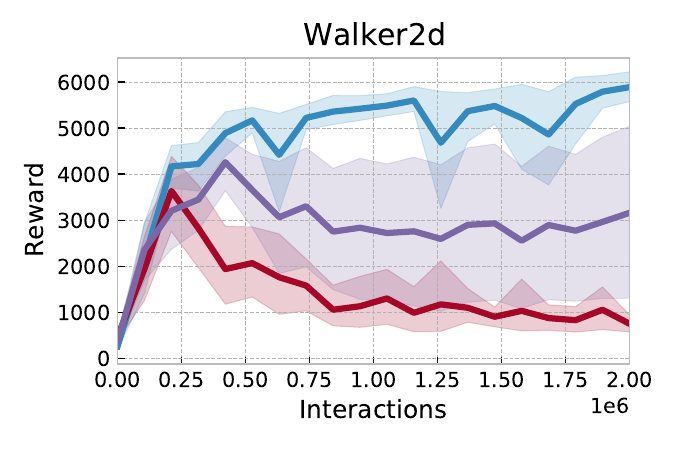}
  \includegraphics[width=0.32\textwidth]{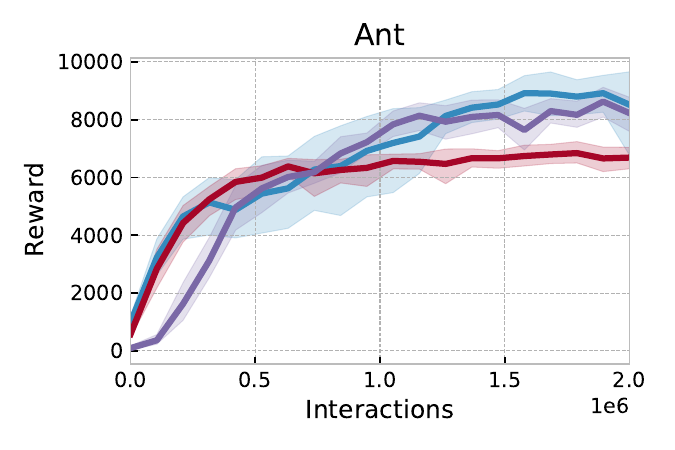} \\
  \includegraphics[width=0.32\textwidth]{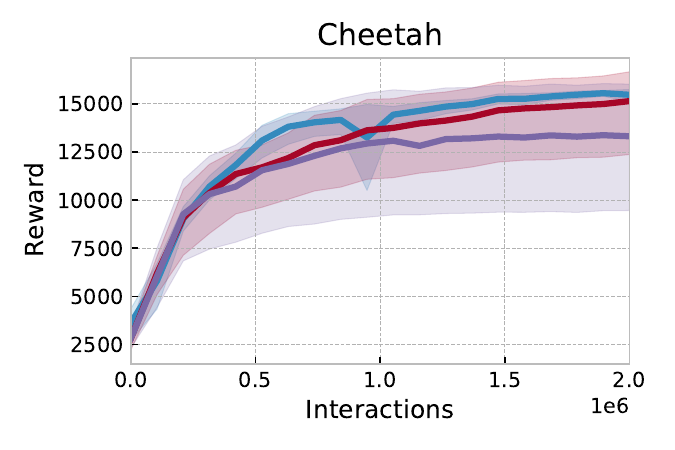}
  \includegraphics[width=0.32\textwidth]{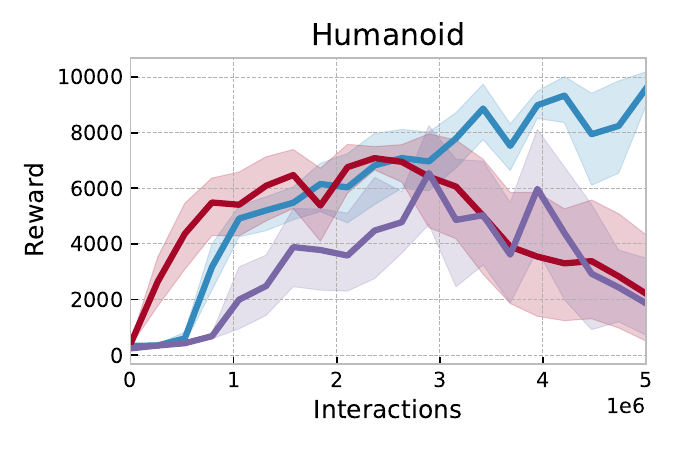}
  \begin{tikzpicture}[every node/.style = {anchor=west,font=\footnotesize}]
    \node at (0,0) {\hspace{0.1mm}\cblock{52}{138}{189}\hspace{2.5mm} Actor \svgH~\thispaper};
    \node at (0,-3.7mm) {\hspace{0.1mm}\cblock{166}{6}{40}\hspace{2.5mm} Critic MVE};
    \node at (0,-8mm) {\hspace{0.1mm}\cblock{128}{114}{179}\hspace{2.5mm} Critic MVE + Actor \svgH};
    \node at (0,-18mm) {}; %
  \end{tikzpicture}\vspace{-4mm}
  \caption{We ablate model rollouts on critic updates
    and actor updates. Since value learning is particularly sensitive to
    dynamics model error, on more complex tasks a slightly inaccurate
    model can halt MVE agent improvement.}
\label{fig:mve_ablation}
\end{figure*}
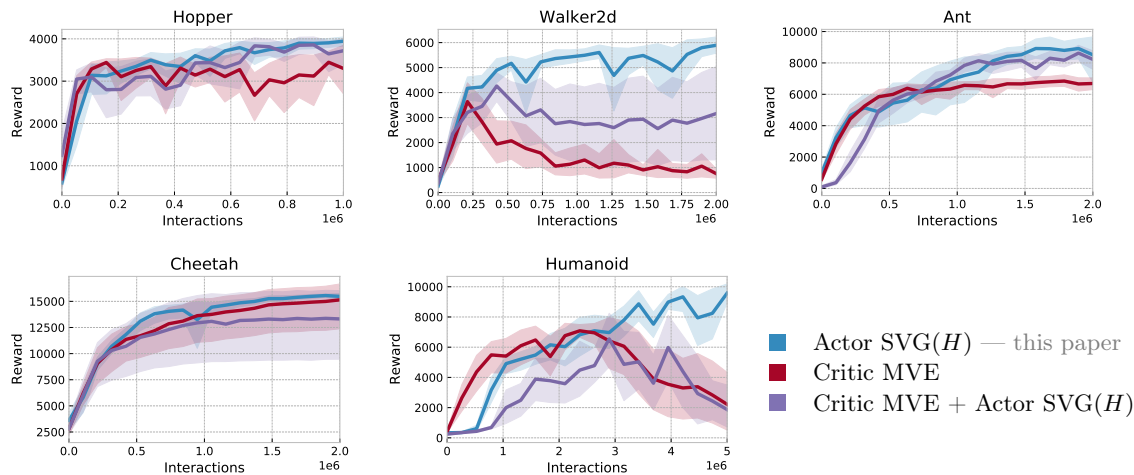

In \cref{fig:mve_ablation}, we consider the
effect of introducing the critic MVE update of
\citet{feinberg2018model} when combined with either the standard
model-free SAC actor update or \svgH. Of particular interest is the
first combination, since it bears close resemblance to an
ablation performed in \citet{janner2019trust} on the Hopper environment.
In \citet{feinberg2018model}, both the dynamics model and the
actor/critic models are updated with 4 minibatch gradient steps at
each timestep. MBPO periodically trained the dynamics
model to convergence on the full replay buffer, generated a large
batch of fictional transitions, and proceeded to repeatedly update the
actor/critic models on those stored transitions.

The contrast between our results and those of the MBPO ablation are
instructive. Whereas \citet{janner2019trust} reported that MVE
significantly underperformed MBPO in terms of final reward, even when
$H=1$, we find that MVE is competitive on all environments until a
certain point in learning, after which performance gradually
degrades. A likely explanation lies in the MVE dynamics model
update. As the replay buffer grows, the dynamics model update is less
likely to
sample recent transitions in each minibatch, resulting in a gradual
increase in model error (\cref{fig:walker_mve_details}). As in
\cref{fig:dx_architecture_ablation},
the increase in dynamics model error cannot be attributed to the
capacity of the model architecture, since the supervised training
error on an equivalent replay buffer is significantly lower than the
online training error. This observation
highlights the impact of model error when the model-generated
transitions are used to update the critic, and supports
\citet{van2019use} argument that inaccurate
parametric forward dynamics models may be particularly detrimental to
value learning.

\section{Conclusion}
\sacsvgH combines the most effective ideas from recent MBRL research,
resulting in a simple, robust model-based agent.
A few key future directions and applications are in:
\begin{enumerate}
\item \emph{Policy refinement or semi-amortization} as
  in \citet{marino2020iterative}
  interprets the policy as solving a model-based control
  optimization problem and can be combined with differentiable control
  \citep{okada2017path,amos2018dmpc,pereira2018mpc,agrawal2019learning,east2020infinite}.
  Fine-tuning can also be performed at an episode-level as
  in \citet{nagabandi2018learning}
  to adapt the dynamics to the observations in the episode.
\item \emph{Constrained MDPs}
  \citep{dalal2018safe,koller2018learning,chow2018lyapunov,dean2019safely,bohez2019value},
  where the model-based value expansion and SVG can guide the agent
  away from undesirable parts of the state space.
\item \emph{Extensions of other model-free algorithms.}
While we only considered SVG extensions to SAC as presented in
\citet{haarnoja2018soft}, similar SVG variations can be added
to the policy learning in other model-free methods such as
\citet{abdolmaleki2018maximum,fujimoto2018addressing,lee2019slac,lee2020sunrise,yarats2019improving}.
\item \emph{Unsupervised pretraining} or \emph{self-supervised learning}
  using world models
  \citep{ha2018worldmodels,shyam2018model,pathak2019self,sharma2020dynamics,sekar2020planning}
  push against the \emph{tabula rasa}
  viewpoint of agents starting from zero knowledge
  about the environment it is interacting with
  and would allow them to start with a reasonable idea
  of what primitive skills the system enables them
  to do.
\item \emph{Going beyond the single-agent, single-task, online setting.}
  Some of the core ideas behind model-based value expansion can
  be applied to more sophisticated settings. In multi-agent
  settings, an agent can consider short-horizon rollouts
  of the opponents. In the batch RL setting
  \citep{fujimoto2018bcq,kumar2019stabilizing,wu2019behavior}
  the short-horizon rollouts can be used to constrain
  the agent to only considering policies that keep the
  observations close to the data manifold of observed trajectories.
\end{enumerate}

\acks{
  We thank Alex Terenin, Maximilian Nickel, and Aravind Rajeswaran
  for insightful discussions
  and acknowledge the Python community
  \citep{van1995python,oliphant2007python}
  for developing
  the core set of tools that enabled this work, including
  PyTorch \citep{paszke2019pytorch},
  Hydra \citep{Yadan2019Hydra},
  Jupyter \citep{kluyver2016jupyter},
  Matplotlib \citep{hunter2007matplotlib},
  seaborn \citep{seaborn},
  numpy \citep{oliphant2006guide,van2011numpy},
  pandas \citep{mckinney2012python}, and
  SciPy \citep{jones2014scipy}.
  This research is supported by an Amazon Research Award, NSF I-DISRE
  193471, NIH R01 DA048764-01A1, NSF IIS-1910266, and NSF 1922658
  NRT-HDR: FUTURE Foundations, Translation, and Responsibility for Data
  Science.
  Samuel Stanton is additionally supported by an
  NDSEG fellowship from the United States Department of Defense.
}

{\footnotesize
\bibliography{svg}
}

\newpage
\appendix
\section{More experimental details}
\label{app:exp}

\begin{algorithm*}[t]
\caption{
  Our combination of SAC and \svgH.
  Components of SAC are in colored in black and
  {\color{sacemph} the model-based SVG components are in purple.}
}
\small
\newcommand{\Dstep}{\gD_{\rm step}}
\newcommand{\Dseq}{\gD_{\rm seq}}
\newcommand{\Mseq}{M_{\rm seq}}
\newcommand{\Mstep}{M_{\rm step}}
\begin{algorithmic}
  \State \textbf{Hyperparameters:}
    \#updates $\Mstep,\Mseq$,
    target network update $\tau$,
    {\color{sacemph}
    planning horizon $H$
    }
  \State \textbf{Models:}
  Actor $\pi_\theta$, critic ensemble $Q_\theta$,
  temperature $\alpha$,
  {\color{sacemph} dynamics $f_\theta$, reward $r_\theta$,
  termination $d_\theta$} \\
  \State Initialize the replay buffer $\gD$
  \For{environment step $t = 1..T$}
    \State Sample $u_t\sim\pi(x_t)$ and execute $u_t$ on the system to obtain
    $(r_t,x_{t+1},d_{t+1})$ and append it to $\gD$
    \For{$\Mstep$ updates}
        \State $\Dstep \leftarrow \{(x_s, u_s, r_s, x_{s+1}, d_{s+1})\}_s\sim_{\rm step}\gD$
        \Comment Sample a batch of single-step transitions
        \State {\color{sacemph} $\theta_\pi \leftarrow \gradupdate(
          \theta_\pi,
            \nabla_{\theta_\pi} \gJ_{\pi,\alpha}^{\rm SVG}(\Dstep)
        )$ \Comment Fit the SVG(H) actor with \eqref{eq:svg_policy}}
        \State $\alpha \leftarrow \gradupdate(
            \alpha,
            \nabla_\alpha \gJ_\alpha(\Dstep)
        )$ \Comment Update the temperature with \eqref{eq:sac:temp}
        \State $\theta_Q \leftarrow \gradupdate(
          \theta_Q,
          \nabla_{\theta_Q} \gJ_Q(\Dstep)
        )$ \Comment Fit the critic ensemble with \eqref{eq:sac:soft_bellman}
        \State {\color{sacemph} $\theta_{r} \leftarrow \gradupdate(
          \theta_{r},
          \nabla_{\theta_{r}} \gJ_r(\Dstep)
          )$
          \Comment Fit the reward model with \eqref{sec:mve_sac:Jr}}
        \State {\color{sacemph} $\theta_{d} \leftarrow \gradupdate(
          \theta_{d},
          \nabla_{\theta_{d}} \gJ_d(\Dstep)
          )$
          \Comment Fit the termination model with \eqref{sec:mve_sac:Jd}}
        \State $\bar\theta_Q \leftarrow \tau \theta_Q + (1-\tau)\bar\theta_Q$
        \Comment Update the target critic ensemble weights
    \EndFor

    {\color{sacemph}
    \For{$\Mseq$ updates}
        \State
        $\Dseq \leftarrow \{x_{s:s+H}\}_s\sim_{\rm seq} \gD$
        \Comment Sample a batch of multi-step transitions
        \State $\theta_f \leftarrow \gradupdate(
          \theta_f,
          \nabla_{\theta_f} \gJ_f(\Dseq)
          )$
          \Comment Fit the multi-step dynamics model with
          \eqref{sec:mve_sac:Jf}
    \EndFor
    }
\EndFor
\end{algorithmic}

\label{alg:sac-svg}
\end{algorithm*}

This section provides more details behind our experiments, including
a time-dependent target entropy in \cref{app:exp:targ_entr},
a description of our hyper-parameters in \cref{app:exp:hypers},
further analysis and descriptions of a walker experiment in \cref{app:exp:walker},
and full plots of our POPLIN experiments in
\cref{fig:exp:poplin_full}.
\Cref{alg:sac-svg} overviews the algorithm describing our combination
of SAC and \svgH.

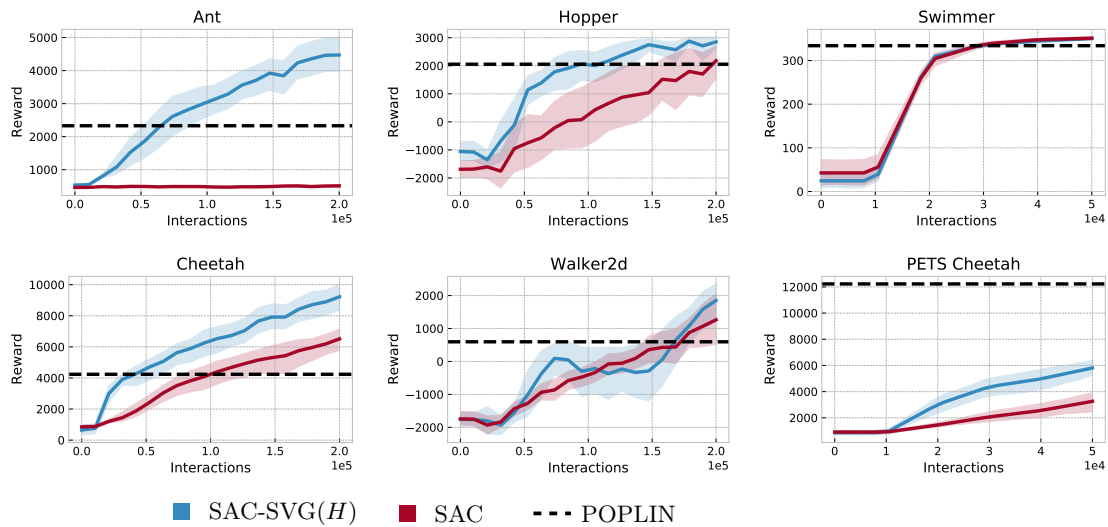
\begin{figure*}[t!]
  \includegraphics[width=0.32\textwidth]{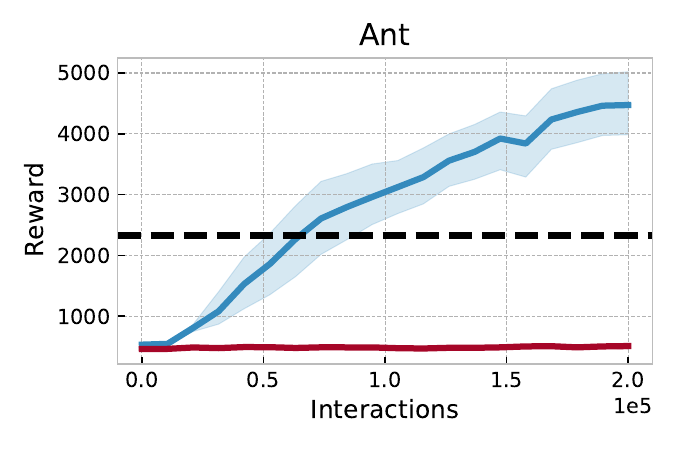}
  \includegraphics[width=0.32\textwidth]{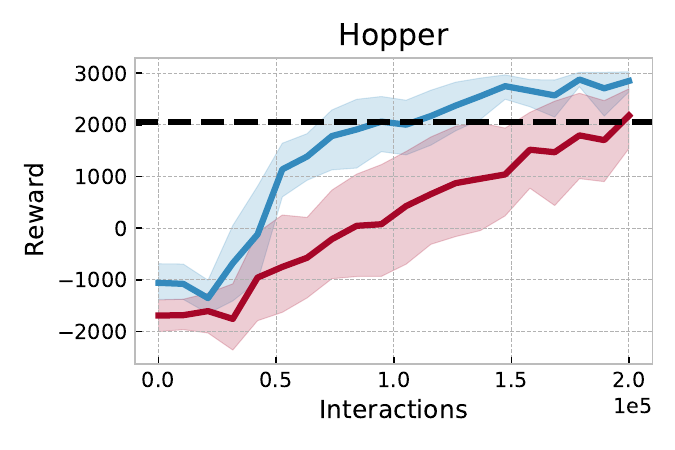}
  \includegraphics[width=0.32\textwidth]{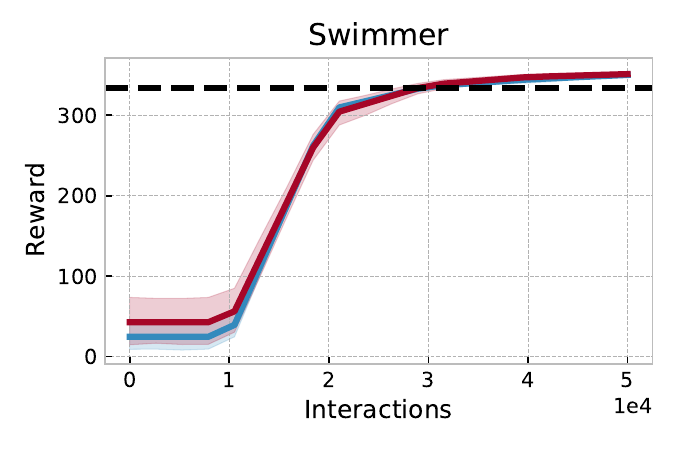} \\
  \includegraphics[width=0.32\textwidth]{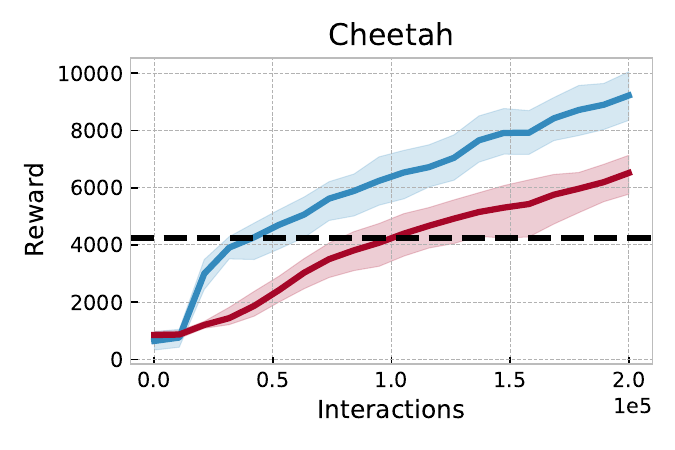}
  \includegraphics[width=0.32\textwidth]{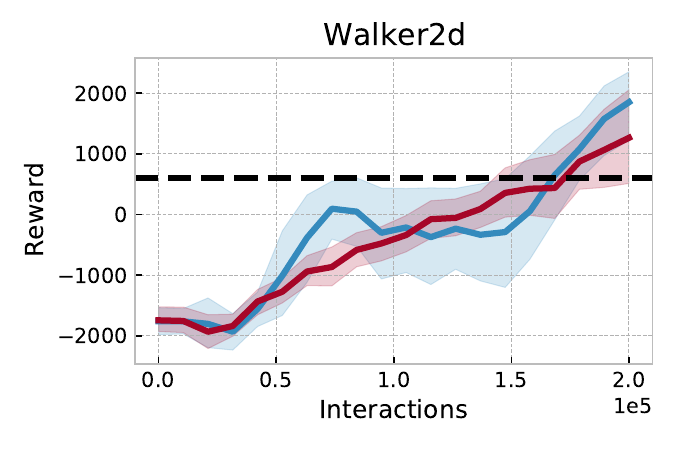}
  \includegraphics[width=0.32\textwidth]{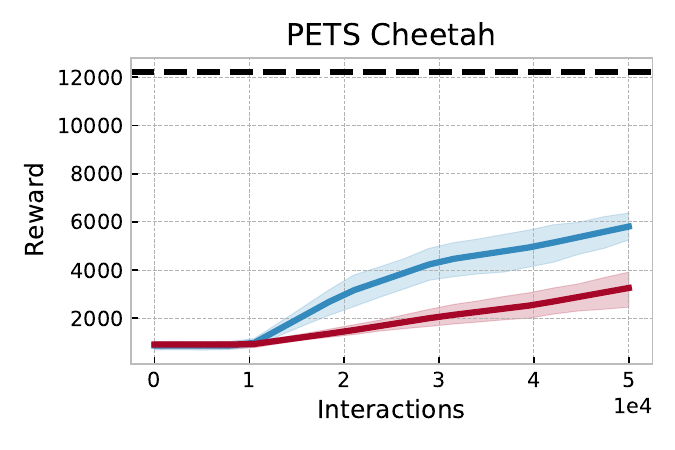} \\
  \hspace*{20mm}
  \begin{tikzpicture}[every node/.style = {anchor=west,font=\footnotesize}]
    \node at (0,0) {\hspace{0.1mm}\cblock{52}{138}{189}\hspace{1.5mm} \sacsvgH};
    \node at (30mm,0mm) {\hspace{0.1mm}\cblock{166}{6}{40}\hspace{1.5mm} SAC};
    \draw [line width=0.5mm,dashed] (50mm,00mm) -- (55mm,0mm);
    \node at (55mm,0mm) {POPLIN};
  \end{tikzpicture}

  \caption{
    Results on the environments tasks considered in POPLIN.
    We run SAC-SVG for ten trials and report the mean and
    standard deviation of the reward.
  }
  \label{fig:exp:poplin_full}
\end{figure*}

\subsection{Time-dependent target entropy}
\label{app:exp:targ_entr}
\begin{wrapfigure}{r}{0.4\textwidth}
  \centering
  \includegraphics[width=0.4\textwidth]{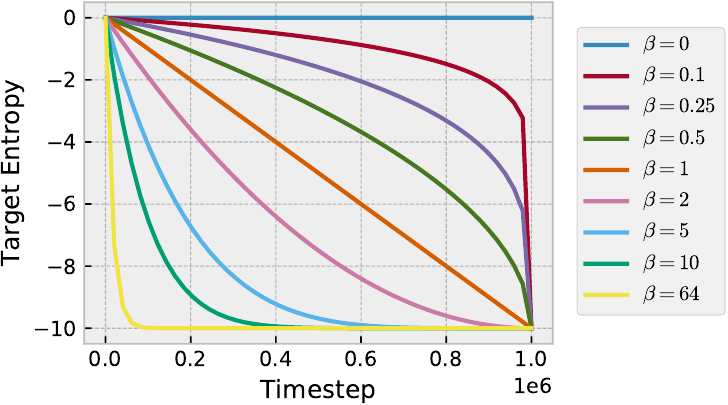}
  \caption{
    Example target entropy decay schedules when training for 1M
    timesteps with a target entropy starting of $0$ and
    ending at $-10$.
  }
  \vspace{-10mm}
  \label{fig:temps}
\end{wrapfigure}
\newcommand{\Ht}{\bar H_t}
\newcommand{\Hinit}{\bar H_\mathrm{init}}
\newcommand{\Hfinal}{\bar H_\mathrm{final}}

We explicitly decay the policy's target entropy $\bar \gH$ rather
than keeping it fixed the entire episode as done in vanilla SAC.
The target entropy is important for balancing exploration and exploitation
and manually decaying it helps control the agent in data-limited settings.
This allows us to start the training with a high-entropy policy
that we explicitly anneal down to a lower entropy by the
end of training. We do this with the exponential decay
\begin{equation}
  \Ht = (\Hinit-\Hfinal)(1-t/T)^\beta + \Hfinal,
  \label{eq:exp_decay}
\end{equation}
where
$T$ is the number of training timesteps,
$\Hinit$ is the initial target entropy,
$\Hfinal$ is the final target entropy,
and $\beta$ is the decay factor.
We plot an example of this in \cref{fig:temps}.

\subsection{Hyper-parameters and random search}
\label{app:exp:hypers}
We share the hyper-parameters in \cref{tab:hypers-shared} between
the tasks and only search over the horizon and target
entropy values, which we show in \cref{tab:hypers}.
We only perform a hyper-parameter search over the
target entropy decay rates for each task, which is
important to learn as it impacts how the agent explores
in the environment. We found \sacsvgH to be more sensitive
to the target entropy decay rate and posit it is important to help
the policy interact with the model-based components in the
earlier phases of training.
We perform a random search over 20 seeds for each task to
find a target entropy schedule, where we sample
$\Hinit\sim {\rm Cat}(\{1,0,-1,-2\})$,
$\Hfinal\sim {\rm Cat}\left(\{\Hinit, -5\} \cup \{-2^i | i\in \{0,\ldots,6\}\}\right)$, and
$\gamma\sim {\rm Cat}\left(\{2^i | i\in\{0,\ldots,6\}\}\right)$,
where ${\rm Cat}(\cdot)$ is a uniform categorical distribution.

\begin{table}[t]
  \caption{Shared hyper-parameters for all tasks.
    SAC's base hyper-parameters are in black and
    {\color{sacemph} our \svgH extensions are in purple.}
  }
  \label{tab:hypers-shared}
  \centering
  \small
  \begin{tabular}{r|l}
    \toprule
    Hyper-Parameter & Value \\ \midrule
    Replay buffer capacity & 1M interactions \\
    All optimizers & Adam \\
    Actor and critic LRs & $10^{-4}$ \\
    Temperature LR & $5\cdot 10^{-4}$ \\
    Init temperature $\alpha_{\rm init}$ & 0.1 \\
    Critic target update rate $\tau$ & $5\cdot 10^{-3}$ \\
    Critic target update freq & every timestep \\
    Actor update freq & every timestep \\
    Discount $\gamma$ & 0.99 \\
    Single-step updates $N_{step}$ & 1 \\
    Single-step batch size & 512 \\
    Actor and critic MLPs & 2 hidden layers, 512 units \\
    Actor log-std bounds & $[-5,2]$ \\
    \midrule
    \color{sacemph} Reward, term, and dx MLPs &
    \color{sacemph} 2 hidden layers, 512 units \\
    \color{sacemph} Dx recurrence &
    \color{sacemph} 2-layer GRU, 512 units \\
    \color{sacemph} Reward, term, and dx LRs &
    \color{sacemph} $10^{-3}$ \\
    \color{sacemph} Multi-step updates $N_{seq}$ &
    \color{sacemph} 4 \\
    \color{sacemph} Multi-step batch size &
    \color{sacemph} 1024 \\ \bottomrule
  \end{tabular}
\end{table}

\begin{table}[t]
  \caption{Task-specific hyper-parameters for the POPLIN (left) and
    MBPO (right) tasks.}
  \label{tab:hypers}
  \vspace{-5mm}
  \begin{minipage}[t]{0.45\linewidth}\vspace{0pt}
  \begin{tabular}{r|ccccc}\toprule
    Environment & $H$ & $\Hinit$ & $\Hfinal$ & $\beta$ \\ \midrule
    Ant & 3 & 1 & -4 & 0.0625 \\
    Hopper & 3 & 1 & 1 & - \\
    Swimmer & 3 & -2 & -16 & 16 \\
    Cheetah & 4 & 0 & -4 & 1 \\
    Walker2d & 5 & 1 & 1 & - \\
    PETS Cheetah & 5 & -2 & -4 & 0.0625 \\\bottomrule
  \end{tabular}
  \vspace{-2mm}
  \end{minipage}
  \hfill
  \begin{minipage}[t]{0.45\linewidth}\vspace{0pt}
  \begin{tabular}{r|cccc}\toprule
    Environment & $\Hinit$ & $\Hfinal$ & $\beta$ \\ \midrule
    Hopper & 0 & -1 & 0.5 \\
    Walker2d & -2 & -3 & 64 \\
    Ant & 2 & -4 & 1 \\
    Cheetah & -2 & -2 & - \\
    Humanoid & -1 & -1 & - \\ \bottomrule
  \end{tabular}
  \end{minipage}
\end{table}

\subsection{Walker experiment analysis when doing critic MVE}
\label{app:exp:walker}

We provide additional data from selected walker trials that use
value expansion on the critic and \svgH expansion on the actor
behind the summary shown in \cref{fig:mve_ablation}.
In MVE trials that perform poorly we
can see the model error increase until eventually the agent stops
improving. As noted in the previous section, this phenomenon
highlights the impact of model error when the model-generated
transitions are used to update the critic, and lends credence to the
argument of \citet{van2019use}, who also suggest that inaccurate
parametric forward dynamics models may be particularly detrimental to
value learning.

\begin{figure}[t]
  \centering
  \hspace*{-20mm}\includegraphics[width=1.25\textwidth]{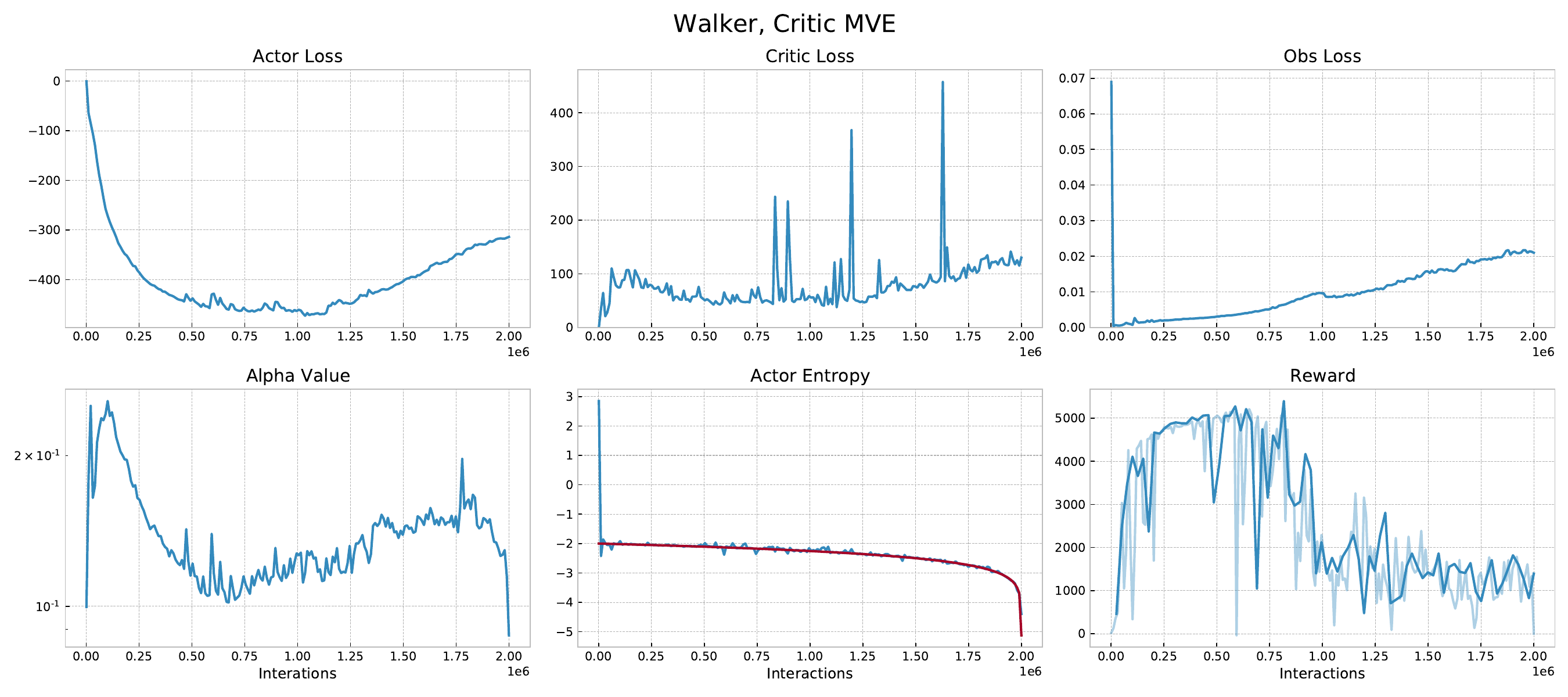} \\[15mm]
  \hspace*{-20mm}\includegraphics[width=1.25\textwidth]{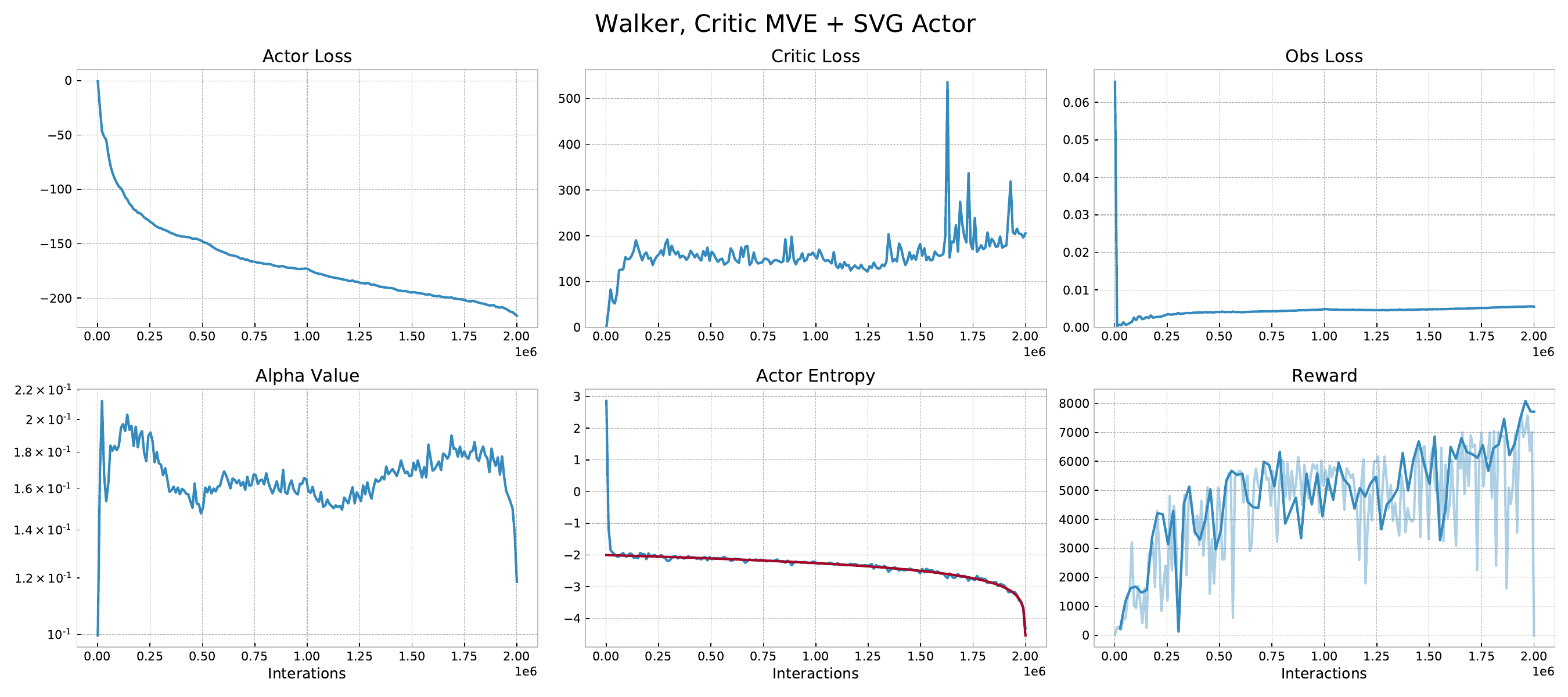}
  \caption{
    The full details behind the Walker runs when performing
    critic value expansion with model-free actor updates (top) and
    critic value expansion with the \svgH actor updates (bottom).
    We find value-expanding in the actor typically leads to more stable behavior.
  }
  \label{fig:walker_mve_details}
\end{figure}

\end{document}